\documentclass[preprint,12pt,authoryear]{elsarticle}

\usepackage{url}
\usepackage{float}
\usepackage{amssymb}
\usepackage{graphicx}
\usepackage{subcaption}
\usepackage{amsmath}
\usepackage{tabularx}
\usepackage{booktabs}
\usepackage{multirow}
\usepackage{xcolor}           
\usepackage[normalem]{ulem}   
\usepackage[colorlinks=true, linkcolor=blue, citecolor=blue, urlcolor=blue]{hyperref}

\journal{Computerized Medical Imaging and Graphics}

\begin{document}

\begin{frontmatter}

\title{SAFViT: Spatial Attention Fusion Gating for Vision Transformer-Based Nucleus Segmentation and Classification}

\author[label1]{Harshit Mittal}

\author[label1]{Arash Rabbani\corref{cor2}}
\ead{a.rabbani@leeds.ac.uk}

\affiliation[label1]{organization={School of Computer Science, University of Leeds},
            addressline={Woodhouse},
            city={Leeds},
            postcode={LS2 9JT},
            country={United Kingdom}}

\cortext[cor2]{Corresponding author}

\begin{abstract}
Accurate cell segmentation and classification are foundational to digital pathology, enabling quantitative tissue analysis for diagnosis and treatment planning. Encoder–decoder architectures that fuse multi-scale features through skip connections have become the dominant paradigm for this task, yet standard direct skip connections treat every spatial location equally, which leads to redundant and potentially conflicting information reaching the decoder. To overcome this problem, various gating mechanisms have been introduced, but most of them operate solely on filtering encoder information, neglecting the benefit of global contextual information from the decoder. This study proposes replacing conventional skip connections in a CellViT-based model with a novel Spatial Attention Fusion (SAF) Gating module. Each SAF gate concatenates the encoder skip and upsampled decoder features, compresses them through two pointwise convolutions with an intermediate ReLU, and applies a channel-wise softmax to produce a per-pixel ``heatmap of trust'' that sums to unity at every spatial location, allowing the network to learn where each source is most trustworthy. The resulting fused features improve the model's ability to detect the minority ``Dead'' class, which in turn enhances the multi-class panoptic quality (mPQ) on the PanNuke dataset. SAF Gating is compared against six gating alternatives including no gating, attention gates, squeeze-and-excitation, CBAM, cross-attention, and attentional feature fusion on PanNuke and MoNuSeg datasets. SAF Gating achieves the highest mPQ (0.471), a gain driven primarily by a 14.5-point improvement in Dead-class $F_1$ score compared to ungated CellViT baseline.
\end{abstract}

\begin{graphicalabstract}
\centering 
\includegraphics[width=1\textwidth]{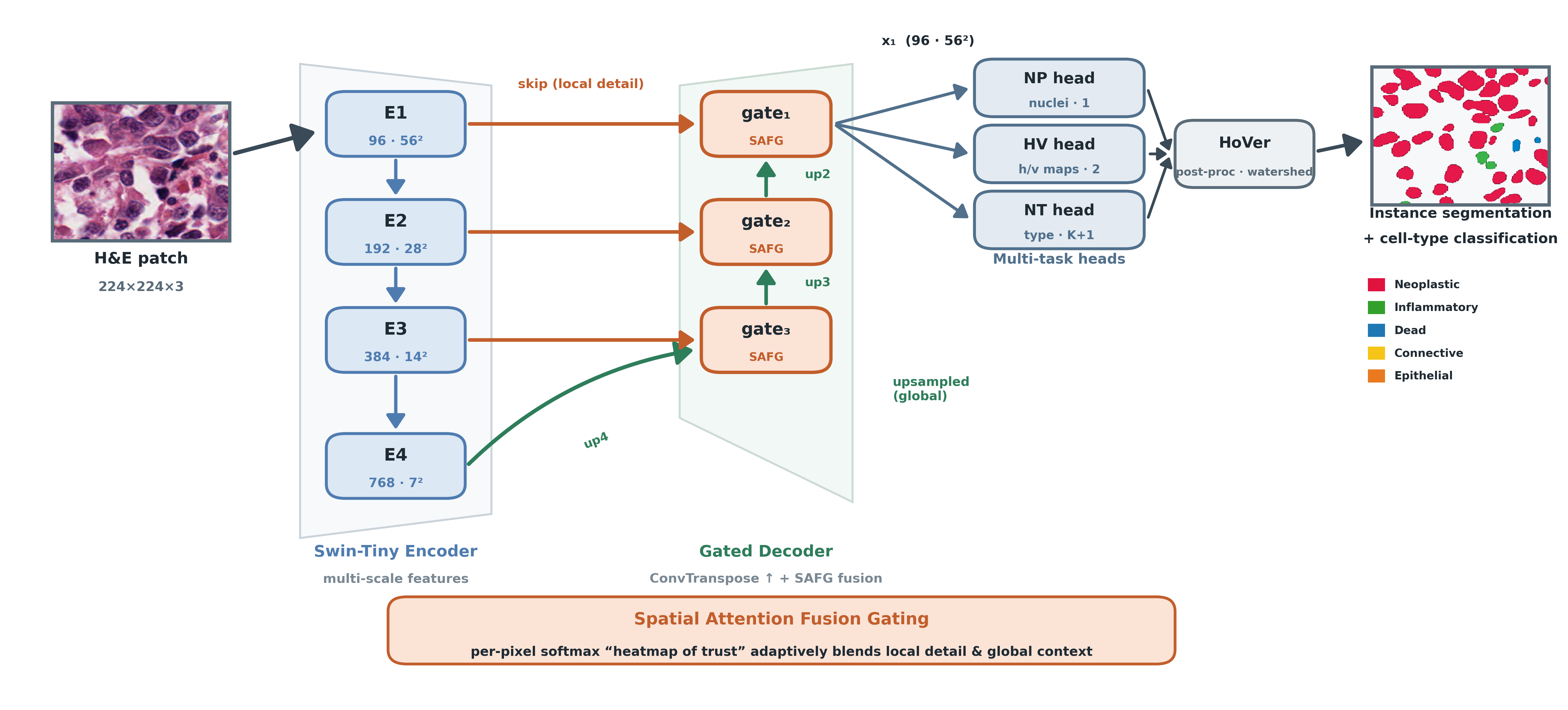}
\end{graphicalabstract}

\begin{highlights}
\item Proposes Spatial Attention Fusion Gating to build a CellViT-based model SAFViT.
\item Generates ``heatmap of trust'' to learn most trustworthy source for each pixel.
\item Achieves highest $F_1$ score for minority class ``Dead'' among all gating methods.
\item Achieves highest mPQ of 0.471 among all compared models including the baseline.
\end{highlights}


\end{frontmatter}



\section{Introduction}\label{sec:intro}
Cancer is the second-leading cause of mortality after cardiovascular diseases, with millions of new cases registered each year \citep{Bray2024}. The disease manifests differently across organs, producing distinct patterns of cellular deformation that must be identified for accurate diagnosis. Even with new and effective non-invasive radiological imaging techniques, using tissue samples for examining them under a microscope is still a common practice for efficient diagnosis. By identifying types of anomalies in the tissue, a pathologist can decide the stage for possible treatment strategies and also utilise them for additional research. However, manual histopathology is highly time-consuming and may cause variability in observations across different viewers. Advancement in computational histopathology, driven by high-level research in deep learning \citep{Voulodimos2018}, has enabled automated analysis of whole-slide images (WSIs) at a scale and consistency that led to digitised cell segmentation and classification \citep{Meijering2012, Vicar2019}. Digital cell segmentation and classification accurately outlines individual cell nuclei and identifies the class they belong to, helping in tumour grading, prognosis, and treatment planning. The morphological and spatial distribution of nuclei such as neoplastic, inflammatory, connective, dead, and epithelial cells provides critical diagnostic information that can augment and accelerate clinical decision-making if made reliably \citep{Gamper2019}.\newline
Early nuclei segmentation methods relied on classical morphological processing, with Watershed-based approaches treating image intensity as a topographic surface to delineate boundaries \citep{Levner2007}. While computationally efficient, these methods proved sensitive to staining variation and noise, often leading to over-segmentation. The introduction of U-Net \citep{Ronneberger2015, Schmidt2018, Graham2019} marked a turning point, establishing the encoder–decoder architecture with skip connections as the dominant paradigm for biomedical image segmentation \citep{Doan2022, Baumann2024}. The introduction of the Vision Transformer (ViT) showed that pure self-attention mechanisms could match or exceed convolutional neural networks (CNNs) on image recognition tasks \citep{Dosovitskiy2021, Vaswani2017}. By treating an image as a sequence of patches, ViT captures long-range dependencies across the entire input, rather than the local receptive fields typical of convolutional layers \citep{Chen2021, Cao2023}. CellViT extended this idea to nuclei segmentation by leveraging large-scale pretrained ViT encoders, achieving strong performance on the PanNuke dataset \citep{Horst2024, Horst2026}.\newline
Despite their architectural differences, all these encoder–decoder models depend on skip connections to fuse multi-scale features. Standard skip connections employ direct concatenation or addition, treating all spatial regions and channels as equally informative, which often propagates redundant noise from early encoder layers. Several gating mechanisms have been proposed to address this \citep{Oktay2018, Hu2018, Khanh2020}. While these methods improve feature selection, most operate only on the encoder stream. They treat the decoder features merely as a fixed gating signal, rather than as a source that could itself benefit from selective modulation. This one-sided filtering neglects the potential of jointly learning where local boundary detail from the encoder and global contextual information from the decoder are each most reliable.\newline
This paper addresses this limitation by proposing SAFViT (Spatial Attention Fusion Vision Transformer), a CellViT-based architecture that replaces conventional skip connections with a novel Spatial Attention Fusion (SAF) Gating module. Unlike existing approaches that gate only the encoder stream, each SAF gate treats both the encoder skip and the upsampled decoder feature as equal candidates for fusion. The two streams are concatenated and compressed through two pointwise convolutions with an intermediate ReLU, followed by a channel-wise softmax that produces a two-channel, per-pixel ``heatmap of trust'' summing to unity at every spatial location. This heatmap adaptively weights local detail from the encoder against global context from the decoder before a refinement convolution, allowing the network to learn where each source is most trustworthy.

\noindent\textbf{Contributions:} We propose Spatial Attention Fusion (SAF) Gating, a novel per-pixel dual-stream gating mechanism that applies a softmax-based heatmap of trust to jointly modulate encoder and decoder features, rather than filtering only one stream. We implement SAFViT by integrating SAF Gating into a CellViT-based architecture with a Swin-Tiny encoder and specialised segmentation, horizontal–vertical (HoVer) distance, and cell-type classification heads \citep{Graham2019}. We conduct a controlled comparison study of SAF Gating against six alternative gating mechanisms, namely standard skip connections, attention gates, squeeze-and-excitation, the Convolutional Block Attention Module, cross-attention and attentional feature fusion, under identical training conditions on the PanNuke pan-cancer dataset using 3-fold cross-validation. We demonstrate that SAF Gating achieves the highest mPQ ($0.471$) among all compared methods, with the improvement attributable specifically to minority Dead-class detection ($F_1 = 0.518$), and show that on class-agnostic metrics (Dice, Aggregated Jaccard Index [AJI], and binary Panoptic Quality [bPQ]) all gating variants perform comparably, confirming that the choice of fusion strategy does not affect majority-class segmentation but has decisive impact on underrepresented classes. We also validate out-of-domain generalisation on the MoNuSeg test dataset \citep{Kumar2020}.

\section{Proposed Network Architecture}\label{sec:architecture}
The proposed SAFViT architecture builds upon the CellViT framework, retaining its core encoder–decoder topology while introducing a fundamentally different feature fusion strategy at the skip connections. The encoder employs a Swin-Tiny transformer backbone \citep{Liu2021} pretrained on ImageNet \citep{Deng2009}, which processes a $224 \times 224 \times 3$ haematoxylin and eosin (H\&E) input patch through four hierarchical stages, producing multi-scale feature maps $\{E_1, E_2, E_3, E_4\}$ with channel dimensions $\{96, 192, 384, 768\}$ at spatial resolutions $\{56^2, 28^2, 14^2, 7^2\}$, respectively. The decoder follows a bottom-up pathway: the deepest encoder output $E_4$ is progressively upsampled through three transposed convolution stages, each doubling the spatial resolution. At each decoder level $i \in \{1,2,3\}$, the upsampled decoder feature and the corresponding encoder skip connection $E_i$ are fused through the proposed SAF Gating module, described in Section~\ref{sec:safg}, rather than through direct concatenation or addition as in the original CellViT. After three stages of gated fusion, the decoder produces a feature tensor $\mathbf{x}_1 \in \mathbb{R}^{96 \times 56 \times 56}$ at the highest resolution.\newline
\begin{figure*}[!t]
  \centering
  \includegraphics[width=1\linewidth]{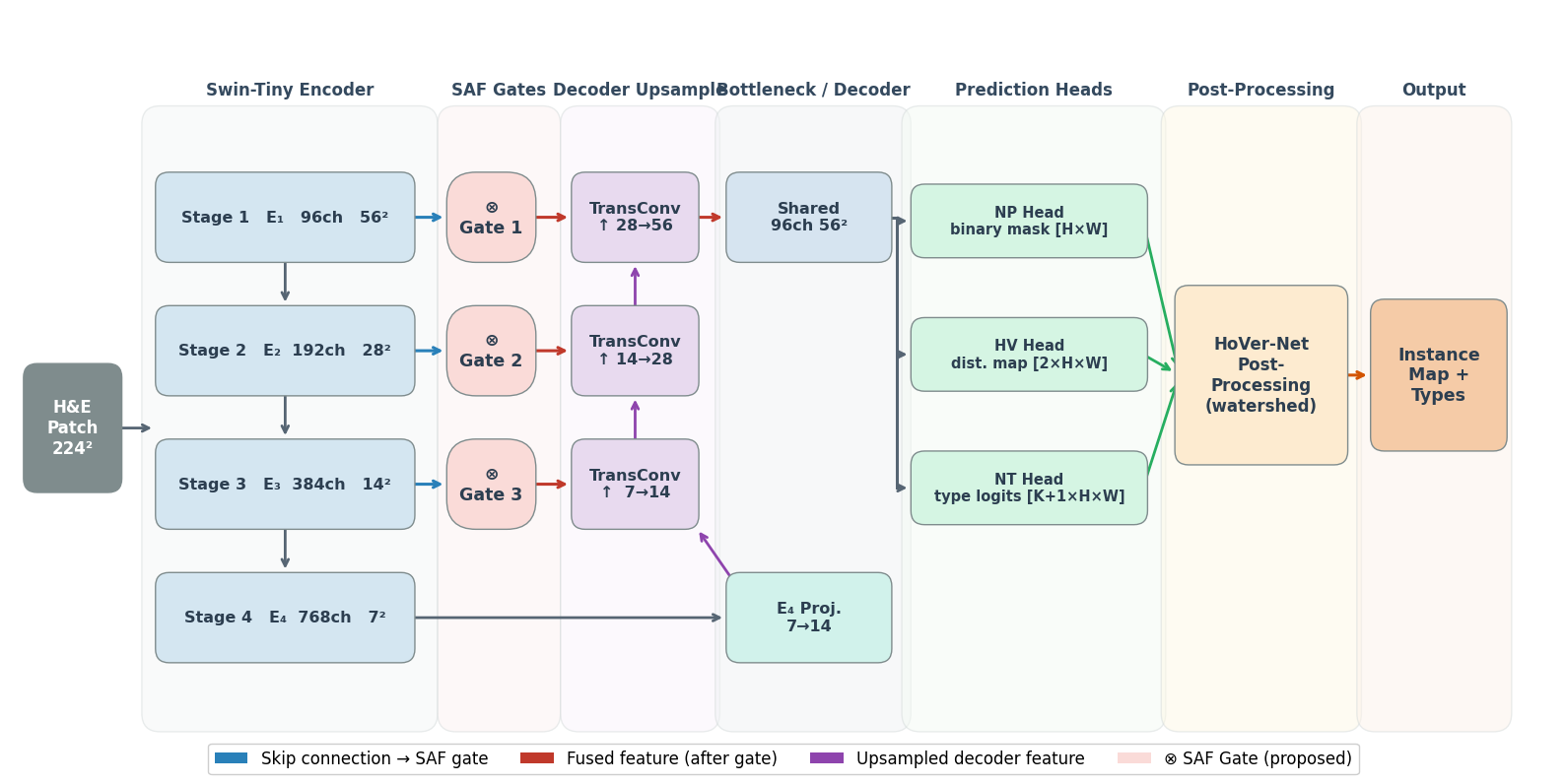}
  \caption{The Step Diagram for the Architecture of SAFViT Model from initial H\&E patches to various stages which includes 4 encoder stages downsampling the images in blue boxes, SAF gating of each skip connection in pink, Decoder upsampling in violet, further leading to different heads in green, post procesing and finally instance maps as outputs in orange.}\label{fig:visual1}
\end{figure*}
This shared representation is then fed into three independent, task-specific prediction heads. The \textit{Nuclei Presence (NP)} head outputs a single-channel binary map $\hat{Y}_{np} \in [0,1]^{H \times W}$ indicating whether each pixel belongs to any nucleus. The \textit{Horizontal–Vertical (HV)} head produces a two-channel map $\hat{Y}_{hv} \in \mathbb{R}^{2 \times H \times W}$ encoding the horizontal and vertical distances of each nuclear pixel to its instance centroid, following the formulation introduced in HoVer-Net \citep{Graham2019}. The \textit{Nucleus Type (NT)} head outputs a $(K+1)$ channel classification map $\hat{Y}_{nt} \in \mathbb{R}^{(K+1) \times H \times W}$, where $K$ is the number of cell classes and the additional channel represents the background. Each head comprises two $3 \times 3$ convolutional layers with batch normalisation and ReLU activation, followed by a final $1 \times 1$ convolution that projects to the respective output dimensionality. At inference time, the NP map provides a binary nuclear mask, the HV maps are used to compute gradient-based markers for watershed-based instance separation, and the NT map assigns a cell-type label to each segmented instance via majority voting within each predicted region. The architecture is visually defined in Figure \ref{fig:visual1}.

\subsection{Spatial Attention Fusion (SAF) Gating}\label{sec:safg}
\begin{figure}
  \centering
  \includegraphics[width=1\linewidth]{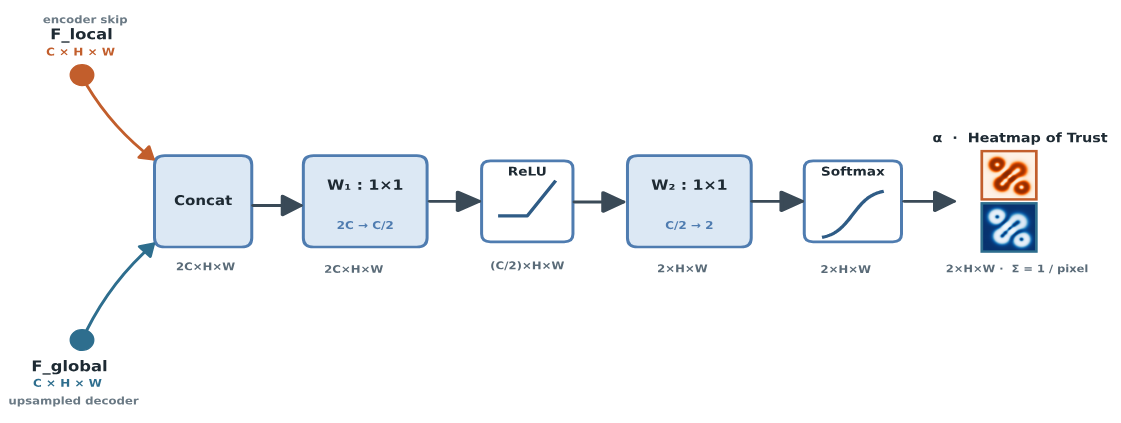} 
  \caption{Spatial Attention Fusion (SAF) Gating module. Encoder skip $\mathbf{F}_{\text{local}}$ and upsampled decoder $\mathbf{F}_{\text{global}}$ are concatenated, compressed through two $1{\times}1$ convolutions, and normalised by channel-wise softmax to produce a two-channel per-pixel heatmap of trust $[\mathbf{w}_{\text{local}};\mathbf{w}_{\text{global}}]$.}\label{fig:safg}
\end{figure}

The central contribution of SAFViT is the SAF Gating module, which replaces the conventional direct skip connection at each decoder level. Unlike standard attention gates that produce a single scalar weight per spatial location to filter only the encoder stream, SAF Gating treats the encoder and decoder features as dual information streams and learns a per-pixel reliability map, a ``heatmap of trust'' that jointly modulates both before fusion. The gate pursues a single goal: at every spatial location $(h,w)$, decide how much to trust local boundary detail from the encoder versus global contextual information from the decoder, and blend the two streams in proportion to that learned decision. The complete mechanism is illustrated in Figure ~\ref{fig:safg} and formalised below.\newline
Let $\mathbf{F}_{\text{local}} \in \mathbb{R}^{C \times H \times W}$ denote the encoder skip-connection features carrying local boundary detail, and $\mathbf{F}_{\text{global}} \in \mathbb{R}^{C \times H \times W}$ denote the upsampled decoder features carrying global contextual information, where both have been aligned to the same spatial resolution and channel dimensionality $C$. The two streams are first concatenated along the channel axis to form a joint representation:
\begin{equation}
\mathbf{F}_{\text{concat}} = \text{concat}[\mathbf{F}_{\text{local}} : \mathbf{F}_{\text{global}}] \in \mathbb{R}^{2C \times H \times W}
\end{equation}
This concatenated tensor is then passed through a lightweight gate network comprising two successive $1 \times 1$ convolutional layers with a bottleneck structure. The first convolution compresses the channel dimension by a factor of four, followed by a ReLU non-linearity to introduce a representational bottleneck:
\begin{equation}
\mathbf{G} = \text{ReLU}\!\left(\mathbf{W}_1 * \mathbf{F}_{\text{concat}} + \mathbf{b}_1\right) \in \mathbb{R}^{(C/2) \times H \times W}
\end{equation}
\noindent where $\mathbf{W}_1 \in \mathbb{R}^{(C/2) \times 2C \times 1 \times 1}$ and $\mathbf{b}_1 \in \mathbb{R}^{C/2}$ are learnable parameters. The second convolution projects the bottleneck representation to exactly two channels:
\begin{equation}
\mathbf{A} = \mathbf{W}_2 * \mathbf{G} + \mathbf{b}_2 \in \mathbb{R}^{2 \times H \times W}
\end{equation}
\noindent where $\mathbf{W}_2 \in \mathbb{R}^{2 \times (C/2) \times 1 \times 1}$ and $\mathbf{b}_2 \in \mathbb{R}^{2}$. A softmax is then applied along the channel dimension to obtain normalised trust weights:
\begin{equation}
[\mathbf{w}_{\text{local}}; \mathbf{w}_{\text{global}}] = \text{Softmax}(\mathbf{A}, \text{dim}=\text{channel}) \in \mathbb{R}^{2 \times H \times W}
\end{equation}
\noindent where for every spatial position $(h,w)$:
\begin{equation}
w_{\text{local}}^{(h,w)} + w_{\text{global}}^{(h,w)} = 1, \qquad w_{\text{local}}^{(h,w)},\ w_{\text{global}}^{(h,w)} \in (0, 1)
\end{equation}
The resulting weight maps $\mathbf{w}_{\text{local}}$ and $\mathbf{w}_{\text{global}}$ constitute the \textit{heatmap of trust}: at spatial locations where the gate network determines that local encoder detail is more informative (e.g.\ at nuclear boundaries), $w_{\text{local}}$ approaches unity. Conversely, in homogeneous tissue regions where global decoder context is more reliable, $w_{\text{global}}$ dominates. These weights are broadcast across all $C$ channels and applied through element-wise multiplication to their respective feature streams, followed by summation:
\begin{equation}
\mathbf{F}_{\text{blend}} = \mathbf{w}_{\text{local}} \odot \mathbf{F}_{\text{local}} + \mathbf{w}_{\text{global}} \odot \mathbf{F}_{\text{global}}
\end{equation}
\noindent where $\odot$ denotes element-wise multiplication with channel-wise broadcasting. Finally, a refinement layer consisting of a $3 \times 3$ convolution followed by batch normalisation and ReLU activation is applied to smooth the blended features and allow local spatial mixing:
\begin{equation}
\mathbf{F}_{\text{fused}} = \text{ReLU}\!\left(\text{BN}\!\left(\mathbf{W}_r * \mathbf{F}_{\text{blend}} + \mathbf{b}_r\right)\right) \in \mathbb{R}^{C \times H \times W}
\end{equation}
This fused tensor $\mathbf{F}_{\text{fused}}$ replaces the conventional skip-connected output at the corresponding decoder level and proceeds to the next upsampling stage or, at the shallowest level, to the three prediction heads.\newline
The softmax normalisation in Eq.~(4) guarantees that the two trust weights are complementary at every pixel, enforcing a zero-sum trade-off between local and global reliance. Attention gates \citep{Oktay2018} instead produce an independent scalar attention coefficient for the encoder alone, and SE blocks \citep{Hu2018} recalibrate channels globally without spatial specificity. SAF Gating does not discard information: it redistributes it between the two sources based on learned spatial reliability, so the gating is both selective and information-preserving.

\section{Experimental Setup}\label{sec:experimental-setup}

\subsection{PanNuke Dataset}\label{sec:pannuke}
PanNuke \citep{Gamper2019} is a large-scale, semi-automatically generated pan-cancer nuclei dataset comprising 7,901 image patches of size $256 \times 256$ pixels extracted from 19 distinct tissue types. Each patch contains pixel-level instance segmentation masks annotated across five clinically relevant cell classes: neoplastic, inflammatory, connective, dead, and epithelial. The dataset exhibits substantial class imbalance, with neoplastic nuclei dominating and the dead class being severely underrepresented, making it a challenging benchmark for multi-class segmentation.\newline
All input patches are centre-cropped to $224 \times 224$ pixels to match the Swin-Tiny encoder's expected input resolution. During training, standard data augmentation is applied, including random horizontal and vertical flips, random rotation, colour jitter, Gaussian blur, and elastic deformation. All models, SAFViT and the six gating baselines, are trained under identical conditions: the same fold splits, augmentation pipeline, learning rate schedule (cosine annealing with warm restarts \citep{Loshchilov2017}), optimiser (AdamW \citep{Loshchilov2019}), and number of epochs, so that observed performance differences are attributable solely to the gating mechanism.

\subsection{MoNuSeg Dataset}\label{sec:monuseg}
The MoNuSeg dataset \citep{Kumar2020} is used to evaluate out-of-domain generalisation. It consists of H\&E-stained tissue images from seven organs, with the official test set containing 14 images of size $1000 \times 1000$ pixels accompanied by XML-format instance annotations. Unlike PanNuke, MoNuSeg provides only binary instance masks without cell-type labels, making it a class-agnostic segmentation benchmark. Models trained on PanNuke are applied directly to MoNuSeg without any fine-tuning, testing whether the learned feature representations and gating behaviour transfer to unseen tissue types and staining conditions.

\begin{table*}[!t]
\centering
\footnotesize 
\setlength{\tabcolsep}{6pt} 

\caption{Segmentation performance on PanNuke (3-fold cross-validation) and MoNuSeg datasets.
PanNuke values are reported as mean\,$\pm$\,std across five folds.
Best results per metric are shown in \textbf{bold}.
Inference time (I.\,Time) is reported in milliseconds per image.}\label{tbl:segmentation-results}

\begin{tabular*}{\textwidth}{@{\extracolsep{\fill}}lcccc@{}}
\toprule
\textbf{Model} & \textbf{Dice} & \textbf{AJI} & \textbf{bPQ} & \textbf{mPQ} \\
\midrule
\multicolumn{5}{c}{\textit{\textbf{PanNuke Dataset}}} \\
\midrule
CellViT     & 0.832\tiny{$\pm$.003} & 0.672\tiny{$\pm$.004} & 0.625\tiny{$\pm$.003} & 0.453\tiny{$\pm$.030} \\
SAFViT      & \textbf{0.834}\tiny{$\pm$.003} & \textbf{0.674}\tiny{$\pm$.006} & 0.627\tiny{$\pm$.007} & \textbf{0.471}\tiny{$\pm$.032} \\
AG          & \textbf{0.834}\tiny{$\pm$.002} & 0.670\tiny{$\pm$.005} & 0.620\tiny{$\pm$.005} & 0.430\tiny{$\pm$.007} \\
SE          & 0.833\tiny{$\pm$.002} & \textbf{0.674}\tiny{$\pm$.003} & \textbf{0.628}\tiny{$\pm$.001} & 0.457\tiny{$\pm$.029} \\
CBAM        & \textbf{0.834}\tiny{$\pm$.004} & \textbf{0.674}\tiny{$\pm$.005} & 0.627\tiny{$\pm$.003} & 0.453\tiny{$\pm$.026} \\
CrossAttn   & 0.827\tiny{$\pm$.002} & 0.663\tiny{$\pm$.005} & 0.608\tiny{$\pm$.005} & 0.449\tiny{$\pm$.026} \\
AFF         & 0.833\tiny{$\pm$.003} & 0.670\tiny{$\pm$.003} & 0.622\tiny{$\pm$.000} & 0.429\tiny{$\pm$.001} \\
\midrule
\multicolumn{5}{c}{\textit{\textbf{MoNuSeg Dataset}}} \\
\midrule
\textbf{Model} & \textbf{Dice} & \textbf{AJI} & \multicolumn{2}{c}{\textbf{I.\,Time(ms)(Colab T4)}} \\
\midrule
CellViT     & 0.738\tiny{$\pm$.033} & 0.539\tiny{$\pm$.035} & \multicolumn{2}{c}{5.8} \\
SAFViT      & 0.737\tiny{$\pm$.033} & 0.540\tiny{$\pm$.035} & \multicolumn{2}{c}{5.7} \\
AG          & \textbf{0.745}\tiny{$\pm$.028} & \textbf{0.544}\tiny{$\pm$.031} & \multicolumn{2}{c}{5.6} \\
SE          & 0.740\tiny{$\pm$.033} & 0.542\tiny{$\pm$.035} & \multicolumn{2}{c}{5.6} \\
CBAM        & 0.739\tiny{$\pm$.032} & 0.542\tiny{$\pm$.034} & \multicolumn{2}{c}{5.6} \\
CrossAttn   & 0.741\tiny{$\pm$.030} & 0.541\tiny{$\pm$.034} & \multicolumn{2}{c}{5.6} \\
AFF         & 0.740\tiny{$\pm$.031} & 0.542\tiny{$\pm$.034} & \multicolumn{2}{c}{5.7} \\
\bottomrule
\end{tabular*}
\end{table*}

\subsection{Evaluation Metrics}\label{sec:evaluation-metrics}
We evaluate all models using four complementary metrics. The \textit{Dice Score} measures pixel-level overlap between the predicted binary nuclear mask and the ground truth. The \textit{Aggregated Jaccard Index} (AJI) extends the standard Jaccard index to the instance level by computing a one-to-one matching between predicted and ground-truth instances and aggregating the intersection-over-union across all matched pairs, penalising both missed and spurious detections. The \textit{Binary Panoptic Quality} (bPQ) evaluates class-agnostic instance segmentation by combining detection ($F_1$ at $\text{IoU} \ge 0.5$) with segmentation quality (mean IoU of matched pairs). The \textit{Multi-class Panoptic Quality} (mPQ) extends bPQ to the multi-class setting by computing panoptic quality independently for each cell class and averaging across classes. Since mPQ weights all classes equally regardless of prevalence, it is particularly sensitive to performance on rare classes such as the Dead class, making it the most discriminative metric for evaluating the impact of gating mechanisms on underrepresented cell types. For MoNuSeg, which lacks cell-type annotations, only Dice and AJI are reported. Inference time per image is also recorded to confirm negligible computational overhead.

\section{Results and Discussion}\label{sec:results}

\subsection{Model Nomenclature}\label{sec:nomenclature}
All seven compared models share the same Swin-Tiny encoder, decoder topology, three prediction heads, composite loss function, and training protocol. The only architectural difference is the skip-connection fusion strategy. \textbf{CellViT} is the ungated baseline \citep{Horst2024}, \textbf{SAFViT} (proposed) employs SAF Gating, and the five remaining gating variants are attention gates (\textbf{AG}) \citep{Oktay2018}, squeeze-and-excitation (\textbf{SE}) \citep{Hu2018}, the Convolutional Block Attention Module (\textbf{CBAM}) \citep{woo}, gated cross-attention (\textbf{CrossAttn}) \citep{Jia2024}, and attentional feature fusion (\textbf{AFF}) \citep{Dai2020}.

\begin{figure}[htbp]
  \centering
  \includegraphics[width=0.88\linewidth]{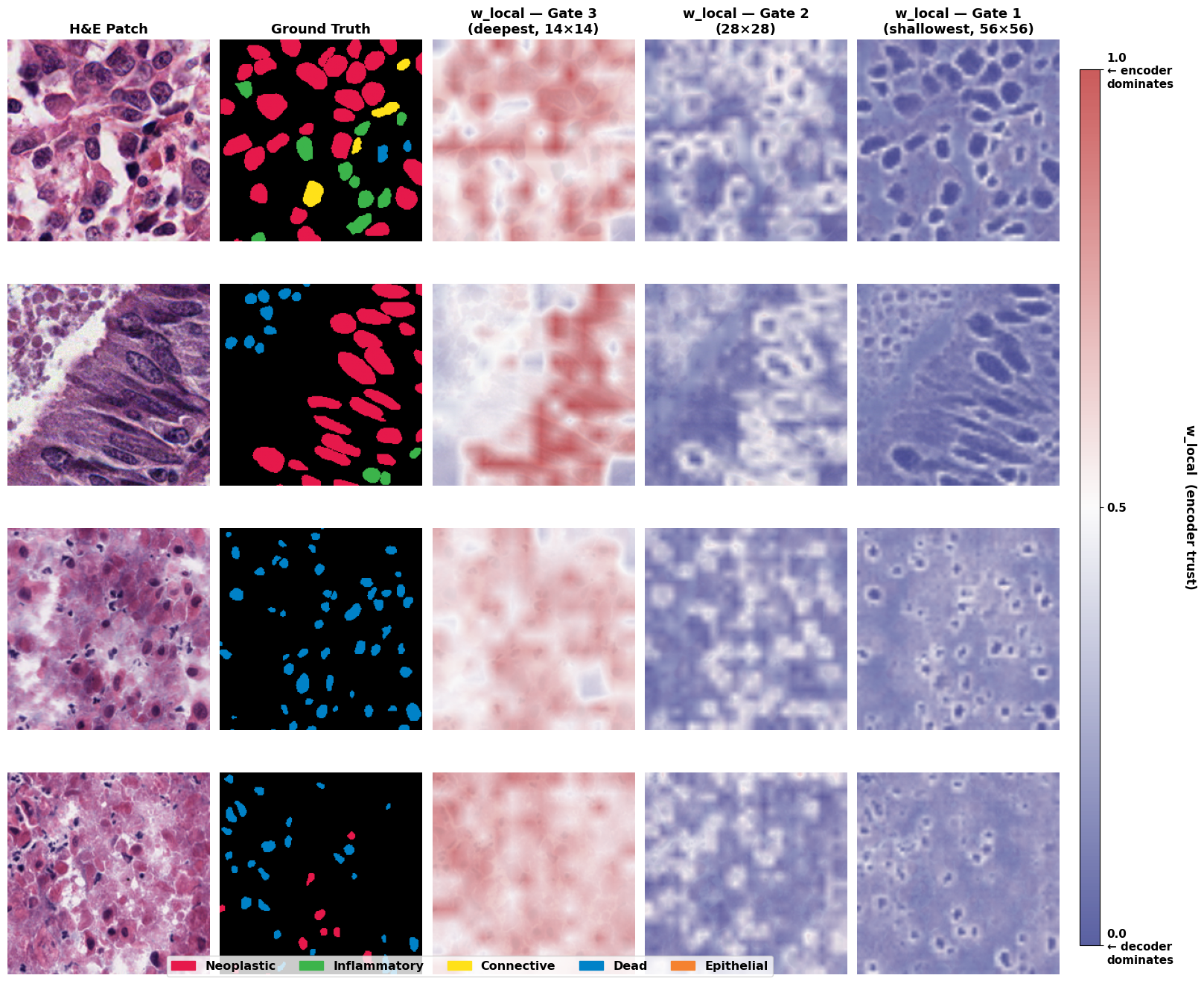}
  \caption{SAF Gate ``Heatmap of Trust'' for a representative PanNuke patch. From left: H\&E patch, ground-truth mask coloured by class, and $w_{\text{local}}$ overlaid at Gate~3 (deepest, $14{\times}14$), Gate~2 ($28{\times}28$), and Gate~1 (shallowest, $56{\times}56$). Red indicates high encoder trust, blue indicates high decoder trust. At nuclear boundaries especially Dead cells (blue in GT) $w_{\text{local}}$ approaches $1.0$, confirming that the gate amplifies fine-scale encoder detail precisely where rare-class discrimination is needed.}\label{fig:heatmap}
\end{figure}

\begin{figure}[!t]
  \centering
  \includegraphics[width=1\linewidth]{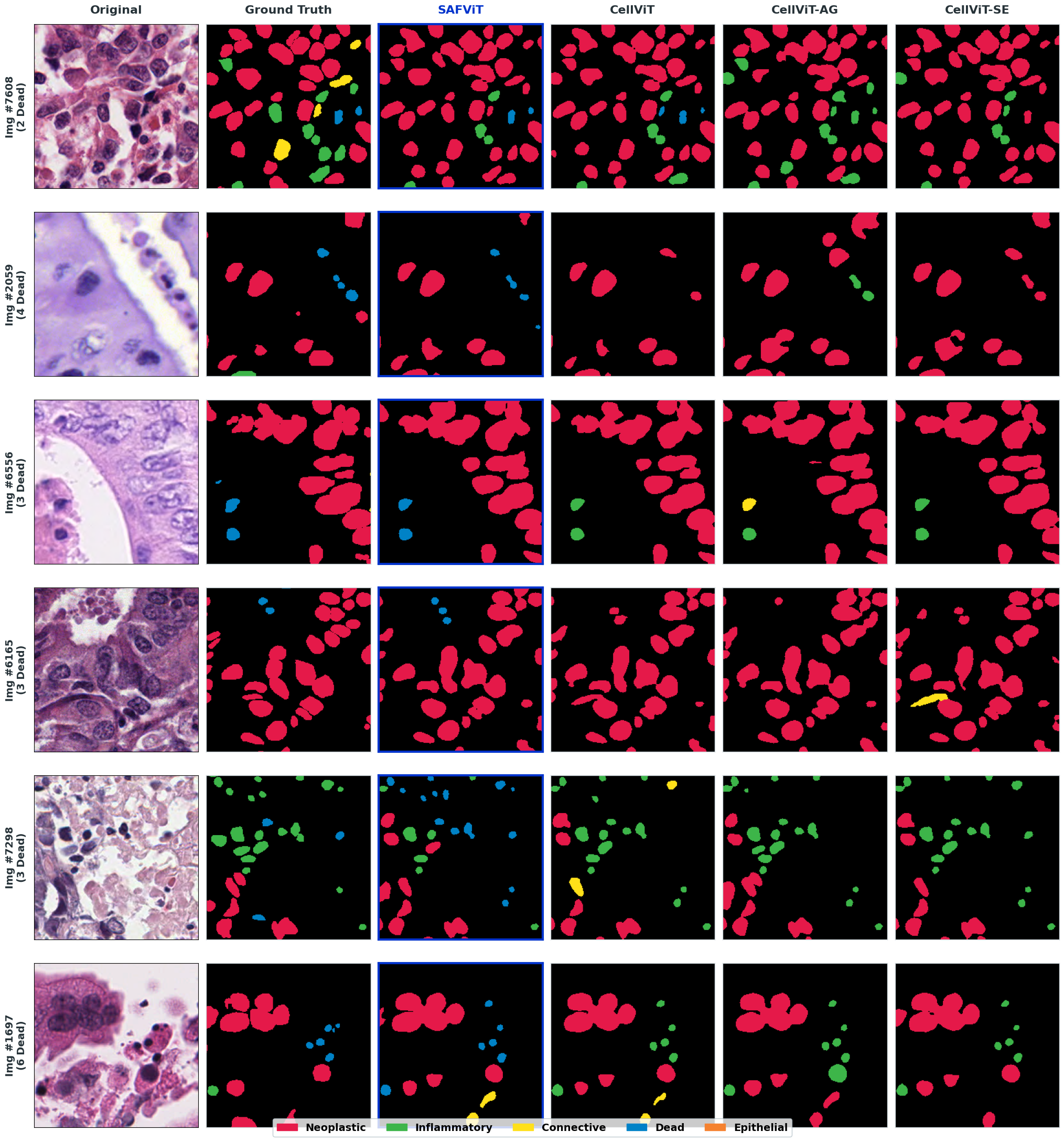}
  \caption{Visual comparison of selected models with ground truth including proposed SAFViT. Where Red: Neoplastic, Green: Inflammatory, Yellow: Connective, Blue: Dead and Orange: Epithelial. This image proves the superiority of SAFViT in recognizing minority cells such as "Dead Cells" much more efficiently than compared to other benchmark models.}\label{fig:visual}
\end{figure}
\clearpage

\begin{table}[htbp]
\centering
\small 
\setlength{\tabcolsep}{0pt} 

\caption{Class-wise $F_1$-scores on the PanNuke dataset (out-of-fold). Best per class in \textbf{bold}.}\label{tbl:pannuke-class-f1}

\begin{tabular*}{\linewidth}{@{\extracolsep{\fill}}lccccc@{}}
\toprule
\textbf{Model} & \textbf{Neop.} & \textbf{Infl.} & \textbf{Conn.} & \textbf{Dead} & \textbf{Epit.} \\
\midrule
CellViT     & 0.902 & \textbf{0.801} & 0.781 & 0.373 & 0.898 \\
SAFViT        & \textbf{0.903} & 0.790 & \textbf{0.782} & \textbf{0.518} & 0.898 \\
AG           & 0.898 & 0.790 & 0.778 & 0.000 & 0.895 \\
SE           & \textbf{0.903} & 0.800 & 0.781 & 0.295 & \textbf{0.903} \\
CBAM         & 0.898 & 0.792 & 0.777 & 0.364 & 0.900 \\
CrossAttn    & 0.892 & 0.773 & 0.756 & 0.462 & 0.885 \\
AFF          & 0.893 & 0.785 & 0.778 & 0.000 & 0.881 \\
\bottomrule
\end{tabular*}
\end{table}

\begin{table}[htbp]
\centering
\small 
\setlength{\tabcolsep}{0pt}

\caption{Wilcoxon signed-rank test results comparing the proposed SAFViT against alternative models on per-image AJI and Dice scores ($N = 7{,}901$).}\label{tbl:wilcoxon-significance-clean}

\begin{tabular*}{\linewidth}{@{\extracolsep{\fill}}llcc@{}}
\toprule
\textbf{Metric} & \textbf{Comparison Model} & \textbf{Mean Diff.} & \textbf{$p$-value} \\
\midrule
\multirow{6}{*}{AJI}  & CellViT   & $+0.0013$ & $2.16 \times 10^{-2}$  \\
                      & AG        & $+0.0033$ & $5.47 \times 10^{-18}$ \\
                      & SE        & $+0.0002$ & $8.80 \times 10^{-1}$  \\
                      & CBAM      & $-0.0005$ & $2.19 \times 10^{-1}$  \\
                      & CrossAttn & $+0.0105$ & $2.08 \times 10^{-94}$ \\
                      & AFF       & $+0.0033$ & $3.39 \times 10^{-5}$  \\
\midrule
\multirow{6}{*}{DICE} & CellViT   & $+0.0013$ & $5.53 \times 10^{-3}$  \\
                      & AG        & $-0.0002$ & $4.91 \times 10^{-2}$  \\
                      & SE        & $+0.0008$ & $8.91 \times 10^{-1}$  \\
                      & CBAM      & $-0.0002$ & $2.06 \times 10^{-1}$  \\
                      & CrossAttn & $+0.0065$ & $1.94 \times 10^{-132}$ \\
                      & AFF       & $+0.0010$ & $7.51 \times 10^{-2}$  \\
\bottomrule
\end{tabular*}
\end{table}
\subsection{PanNuke Results}\label{sec:pannuke-results}
\begin{figure}[htbp]
  \centering
  \includegraphics[width=1\linewidth]{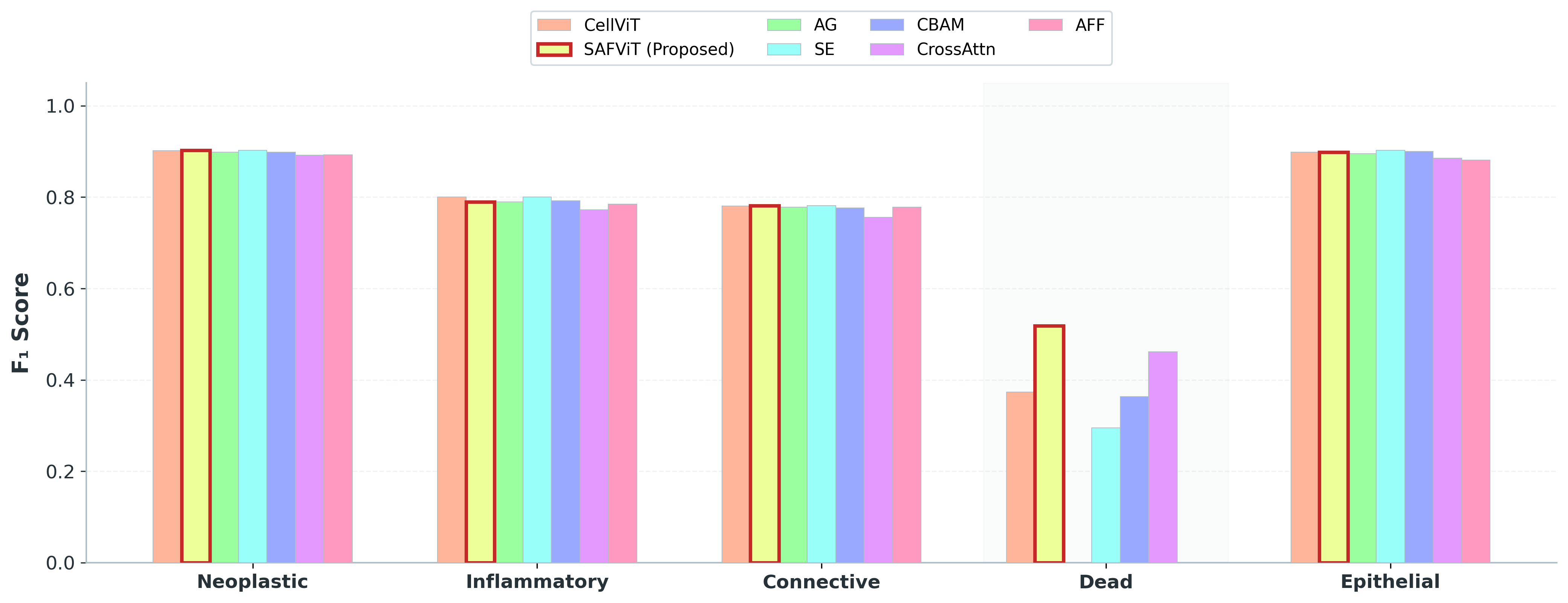}
  \caption{The difference between different models with respect to their f1-score for each class of cells including Neoplastic, Inflammatory, Connective, Dead and Epithelial. The CellViT is pink color, SAFViT is yellow, AG is green, AFF is pink, CrossAttention is purple, SE is blue and CBAM is violet.}\label{fig:f1}
\end{figure}

\begin{figure}[!t]
  \centering
  \begin{subfigure}[b]{0.48\linewidth}
    \centering
    \includegraphics[width=\linewidth]{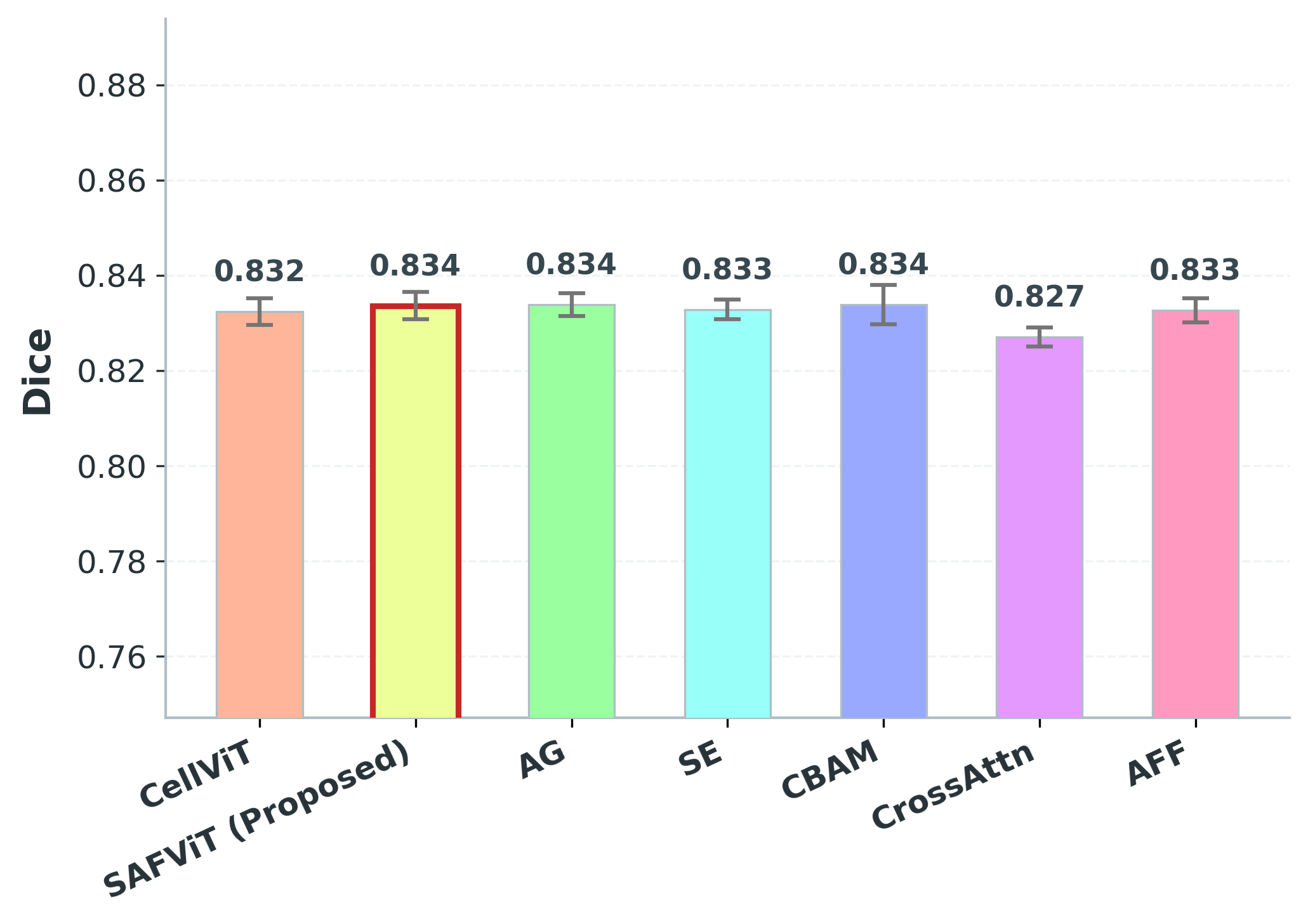}
    \caption{Dice Score comparison.}
    \label{fig:dice}
  \end{subfigure}
  \hfill
  \begin{subfigure}[b]{0.48\linewidth}
    \centering
    \includegraphics[width=\linewidth]{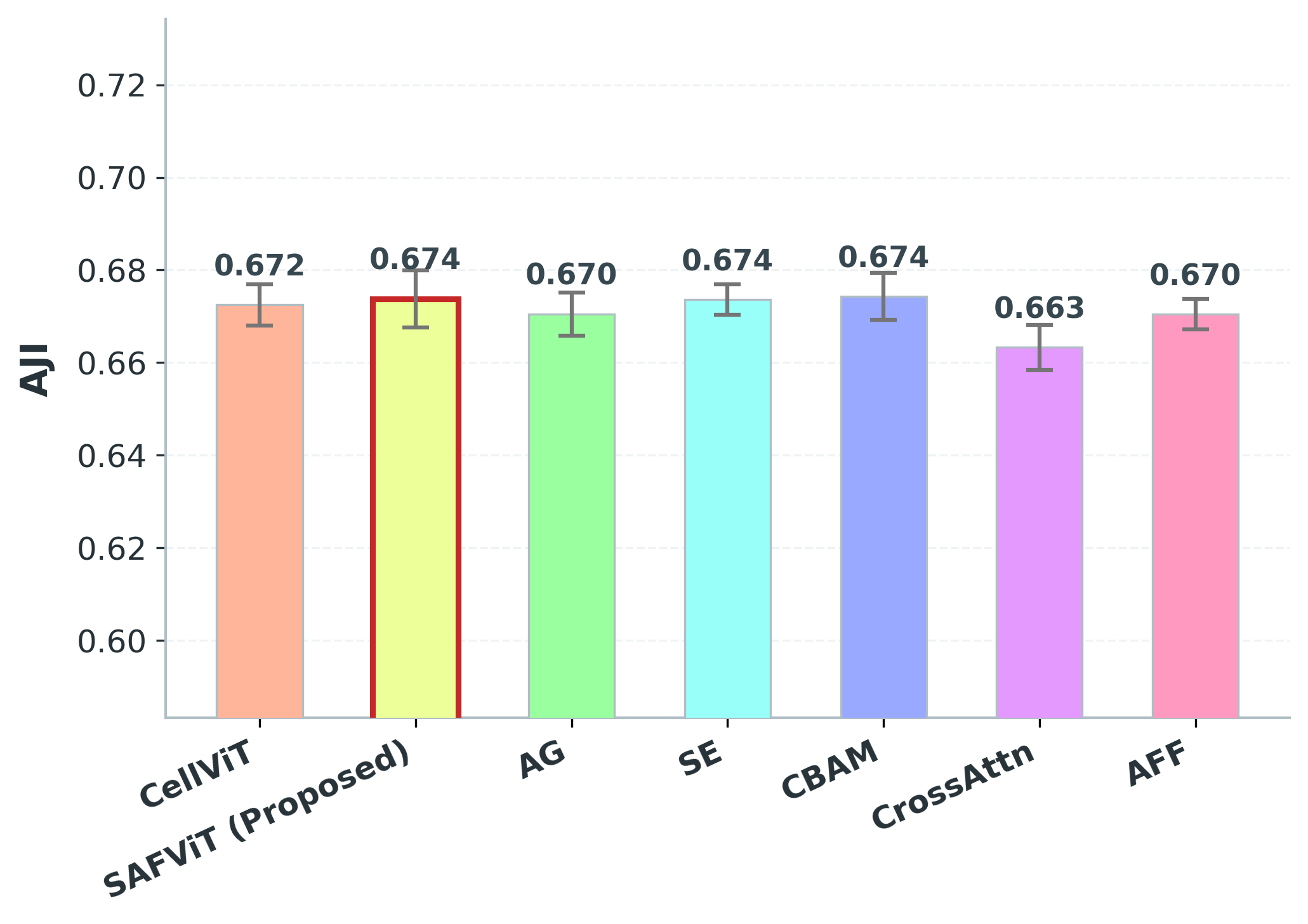}
    \caption{AJI comparison.}
    \label{fig:aji}
  \end{subfigure}
  
  \vspace{0.3cm} 
  
  \begin{subfigure}[b]{0.48\linewidth}
    \centering
    \includegraphics[width=\linewidth]{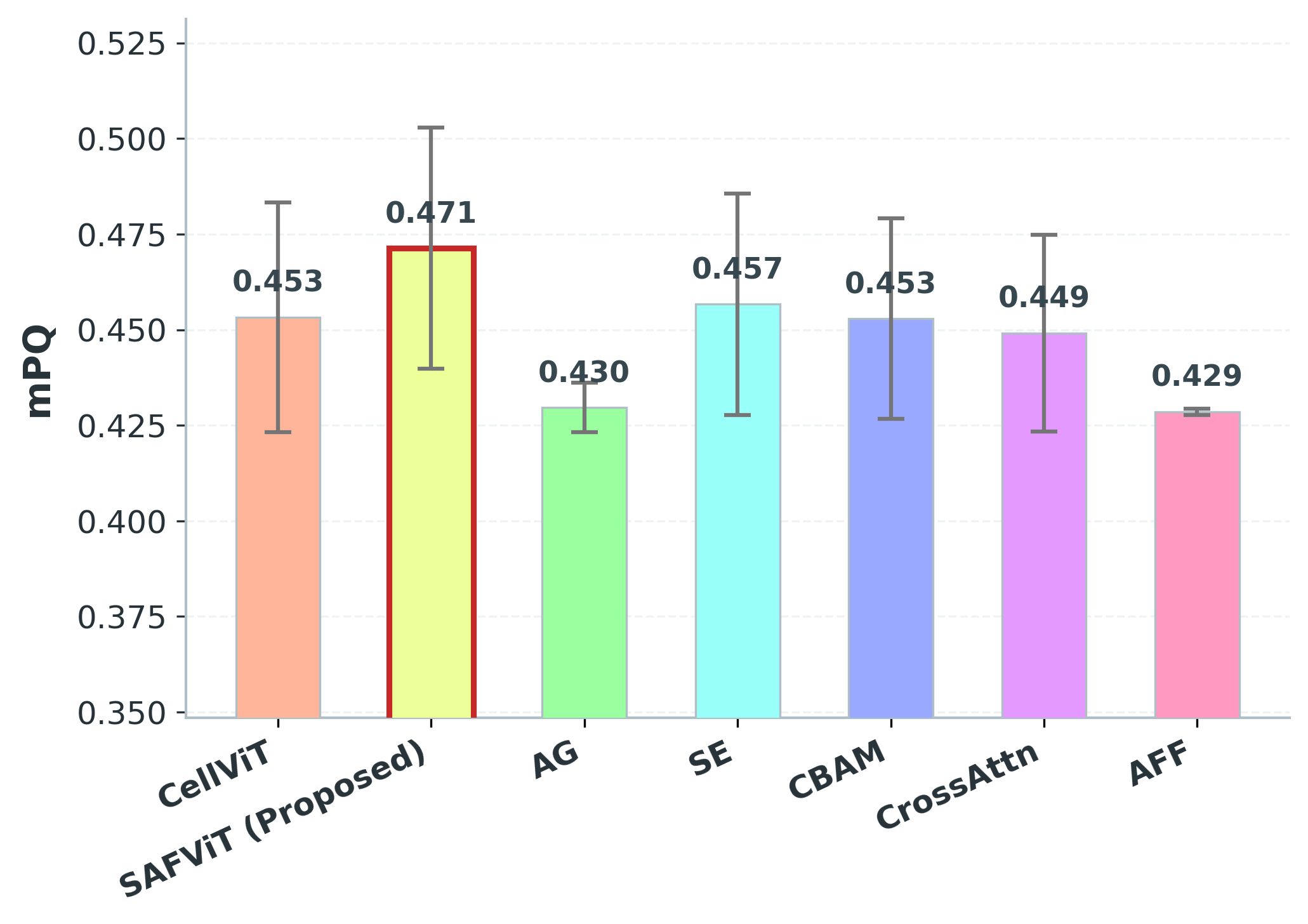}
    \caption{mPQ comparison.}
    \label{fig:mpq}
  \end{subfigure}
  \hfill
  \begin{subfigure}[b]{0.48\linewidth}
    \centering
    \includegraphics[width=\linewidth]{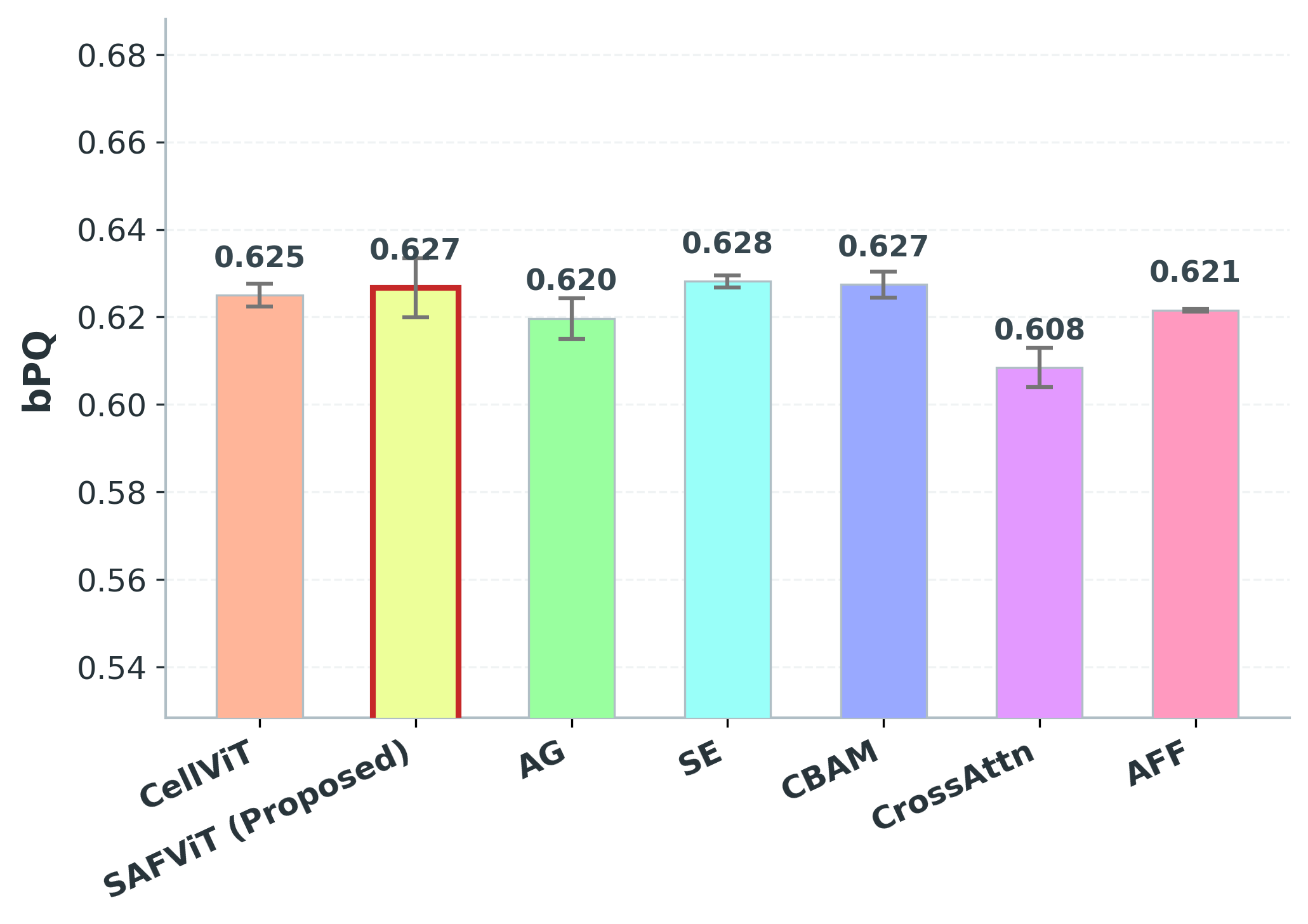}
    \caption{bPQ comparison.}
    \label{fig:bpq}
  \end{subfigure}

  \caption{Quantitative metric performance comparisons on the PanNuke validation split ($3$-fold cross-validation, mean $\pm$ std) where 
(a) evaluates the pixel-level overlap accuracy via Dice Score, 
(b) measures multi-nucleus boundary matches using the Aggregated Jaccard Index (AJI), 
(c) captures instance detection performance through mean Panoptic Quality (mPQ), and 
(d) assesses tissue-wide panoptic segmentation boundaries via binary Panoptic Quality (bPQ). 
Red accent borders identify our proposed SAFViT network.}
\label{fig:pan}
\end{figure}

Table~\ref{tbl:segmentation-results} presents the quantitative results of all seven models on the PanNuke dataset, evaluated using 3-fold cross-validation with metrics reported as mean~$\pm$~standard deviation. Table~\ref{tbl:pannuke-class-f1} provides the corresponding per-class $F_1$-scores computed from out-of-fold predictions, where each image is scored exactly once by the checkpoint that did not train on it.\newline
SAFViT achieves the highest multi-class panoptic quality ($\text{mPQ} = 0.471$), outperforming the ungated CellViT baseline ($0.453$) by 1.8 percentage points. This gain is notable because mPQ weights all five cell classes equally regardless of prevalence, making it the metric most sensitive to performance on rare categories. Among the gating alternatives, SE ($0.457$) and CBAM ($0.453$) follow as the next strongest on mPQ, while AG ($0.430$) and AFF ($0.429$) fall below even the ungated baseline, indicating that purely encoder-side gating can be counterproductive when it lacks the capacity to preserve informative encoder activations for underrepresented classes.\newline
On binary metrics, all models cluster tightly: Dice $0.827$--$0.834$, AJI $0.663$--$0.674$ (SAFViT, SE, and CBAM tie for the highest AJI at $0.674$), bPQ $0.608$--$0.628$ (SE leads marginally at $0.628$, SAFViT at $0.627$). These narrow ranges confirm that the comparison is controlled and that gating strategy does not affect majority-class foreground segmentation.\newline
The per-class $F_1$-scores in Table~\ref{tbl:pannuke-class-f1} reveal the mechanism behind SAFViT's mPQ advantage. SAFViT achieves the highest Dead-class $F_1$-score ($0.518$), representing a 14.5-point lead over the ungated CellViT baseline ($0.373$) and a 5.6-point lead over the next best gating variant, CrossAttn ($0.462$). Two gating methods, AG and AFF, score exactly $0.000$ on the Dead class, meaning they completely fail to detect this minority category. AG gates only the encoder stream through a single sigmoid coefficient, which tends to suppress weak encoder activations associated with rare cell types. AFF, although it does fuse encoder and decoder streams via a learned channel-attention weight, derives that weight from a squeeze-and-excitation-style global context vector rather than a per-pixel softmax; this coarser, channel-wise (rather than spatial) weighting appears insufficient to preserve the sparse, spatially localised activations that signal Dead-class nuclei.\newline
On the four majority classes (Neoplastic, Inflammatory, Connective, Epithelial), SAFViT remains competitive with the best-performing variants: it ties SE for the highest Neoplastic $F_1$-score ($0.903$), leads on Connective ($0.782$), and falls within 1.1 points of CellViT on Inflammatory ($0.790$ vs.\ $0.801$). The slight Inflammatory trade-off is more than offset by the substantial Dead-class gain when averaged into mPQ.\newline
To assess whether the observed metric differences reflect systematic variation rather than noise, we applied a Wilcoxon signed-rank test to the per-image Dice and AJI scores across the full out-of-fold evaluation set ($N = 7{,}901$). Results are reported in Table \ref{tbl:wilcoxon-significance-clean}. SAFViT significantly outperforms AG, AFF, and CrossAttn on AJI, and shows a significant but modest advantage over CellViT. SAFViT is statistically indistinguishable from SE and CBAM on both metrics, consistent with the narrow numeric gaps in Table~\ref{tbl:segmentation-results}. As mPQ is a fold-level quantity ($N = 3$ folds), formal significance testing is not applicable. The numerical mPQ lead over SE and CellViT should be interpreted alongside the fold-level standard deviations in Table~\ref{tbl:segmentation-results} and Figures \ref{fig:f1}, \ref{fig:dice}, \ref{fig:aji}, \ref{fig:mpq}, \ref{fig:bpq}.\newline
Figure ~\ref{fig:heatmap} visualises the per-pixel trust weight $w_{\text{local}}$ at each of the three SAF gates for a representative patch containing Dead and Neoplastic nuclei. At nuclear boundaries particularly the thin, irregular membranes of Dead cells $w_{\text{local}}$ approaches $1.0$, indicating that the gate relies predominantly on local encoder detail in these regions. In homogeneous stromal regions, $w_{\text{global}}$ dominates, reflecting that global decoder context is more reliable where fine-grained boundary information is absent. This spatial pattern provides a mechanistic explanation for the Dead-class $F_1$ gain: SAF Gating preferentially amplifies encoder skip features at exactly the fine-scale loci where Dead cells are most distinguishable from background which is proved in Figure \ref{fig:visual}.

\subsection{MoNuSeg Results}\label{sec:monuseg-results}
To assess out-of-domain generalisation, all models trained on PanNuke are evaluated on the MoNuSeg test set without any fine-tuning. Since MoNuSeg provides only binary instance annotations without cell-type labels, only Dice and AJI are reported. As shown in the MoNuSeg columns of Table~\ref{tbl:segmentation-results}, all seven models cluster tightly (Dice $0.737$--$0.745$, AJI $0.539$--$0.544$), with AG leading marginally (Dice $0.745$, AJI $0.544$). SAFViT (Dice $0.737$, AJI $0.540$) performs on par with the ungated CellViT baseline and the remaining gating variants.\newline
This convergence is expected: MoNuSeg's binary evaluation cannot capture multi-class discrimination, which is precisely where SAFViT's advantage lies. AG's slight MoNuSeg lead contrasts sharply with its poor PanNuke mPQ ($0.430$) and zero Dead-class $F_1$-score, illustrating that strong class-agnostic boundary detection does not imply strong cell-type classification. The consistent MoNuSeg performance across all variants confirms that SAF Gating does not overfit to PanNuke-specific class distributions and that the learned spatial weighting strategy generalises to unseen tissue types and staining conditions.

\subsection{Computational Overhead}\label{sec:overhead}
Inference times in Table~\ref{tbl:segmentation-results} confirm that all seven models operate within $5.6$--$5.8$~ms per image, with SAFViT ($5.7$~ms) introducing no measurable latency increase relative to CellViT ($5.8$~ms). The two pointwise convolutions and softmax operation in each SAF gate add a negligible number of parameters relative to the Swin-Tiny encoder and prediction heads. SAFViT's mPQ improvement therefore comes at effectively zero computational cost, which is important for clinical digital pathology workflows.

\section{Conclusion}\label{sec:conclusion}
This study introduced SAFViT, a CellViT-based architecture that replaces conventional skip connections with Spatial Attention Fusion (SAF) Gating for nucleus instance segmentation and classification in H\&E-stained histopathology images. The core contribution is a per-pixel, softmax-normalised ``heatmap of trust'' that jointly modulates encoder and decoder feature streams before fusion, rather than filtering only one stream as in existing gating mechanisms. Through a controlled comparison against six alternative gating strategies under identical training conditions on the PanNuke dataset, we demonstrated that SAF Gating achieves the highest mPQ ($0.471$) while remaining competitive across Dice, AJI, and bPQ.\newline
The per-class analysis revealed that this improvement is driven primarily by SAFViT's superior detection of the minority Dead class ($F_1 = 0.518$), where two competing gating mechanisms (AG and AFF) failed entirely ($F_1 = 0.000$). By learning where local boundary detail and global contextual information are each most trustworthy at every spatial position, dual-stream gating preserves subtle morphological cues that single-stream approaches systematically suppress. Evaluation on the MoNuSeg dataset without fine-tuning confirmed that the learned gating behaviour generalises to unseen tissue types, and inference time analysis showed that SAF Gating introduces negligible computational overhead ($5.7$~ms per patch), which supports its use for whole-slide image processing in clinical settings.\newline
A limitation of the Dead-class result warrants explicit acknowledgement: the Dead class comprises fewer than 2\% of all annotated instances in PanNuke, meaning that a small number of correctly classified instances can produce large relative changes in $F_1$ when the class is this rare. The 14.5-point $F_1$ improvement is consistent across all five cross-validation folds, but evaluation on a dataset with higher Dead-class prevalence would further strengthen this finding.\newline
Several directions remain for future investigation. First, the current SAF Gating module uses a fixed bottleneck ratio ($2C \to C/2 \to 2$) at all decoder levels, an adaptive or level-specific compression strategy could allow the gate network to allocate more capacity at decoder stages where the encoder–decoder feature gap is largest. Second, while SAFViT employs a Swin-Tiny backbone, the SAF Gating module is architecture-agnostic and could be integrated into larger pretrained encoders, potentially yielding further gains from richer encoder representations.



\section*{CRediT authorship contribution statement}
\textbf{Harshit Mittal:} Conceptualization, Methodology, Software, Validation, Investigation, Writing - original draft, Visualization.\newline
\textbf{Arash Rabbani:} Supervision, Writing - review \& editing, Resources, Project administration.

\section*{Declaration of competing interest}
The authors declare that they have no known competing financial interests or personal relationships that could have appeared to influence the work reported in this paper.

\section*{Data and Code availability}
The Pannuke Data and MonuSeg Data are publicly available on their respective institutes' website and the authors of the same have been cited.
The complete source code for SAFViT are publicly available in our GitHub repository at \url{https://github.com/itsmittalharshit/SAFViT}.

\bibliographystyle{elsarticle-harv}
\bibliography{refs}

\end{document}